# A quantitative analysis of tilt in the Café Wall illusion: a bioplausible model for foveal and peripheral vision

Nasim Nematzadeh, David M. W. Powers
School of Computer Science, Engineering and Mathematics
Flinders University
Adelaide, Australia

nasim.nematzadeh, david.powers@flinders.edu.au

*Abstract*— The biological characteristics of human visual processing can be investigated through the study of optical illusions and their perception, giving rise to intuitions that may improve computer vision to match human performance. Geometric illusions are a specific subfamily in which orientations and angles are misperceived. This paper reports quantifiable predictions of the degree of tilt for a typical geometric illusion called Café Wall, in which the mortar between the tiles seems to tilt or bow. Our study employs a common bioplausible model of retinal processing and we further develop an analytic processing pipeline to quantify and thus predict the specific angle of tilt. We further study the effect of resolution and feature size in order to predict the different perceived tilts in different areas of the fovea and periphery, where resolution varies as the eye saccades to different parts of the image. In the experiments, several different minimal portions of the pattern, modeling monocular and binocular foveal views, are investigated across multiple scales, in order to quantify tilts with confidence intervals and explore the difference between local and global tilt.

*Keywords— Visual perception; Bioinspired neural networks; Geometric illusions; Café Wall illusion; Tilt effects; Difference of Gaussian; Pattern recognition*

## I. INTRODUCTION

The study of *human vision* is a multidisciplinary field, connecting the physiology of vision to bioplausible computational modelling, as well as psychophysical experiments in visual psychology. One source of evidence about vision is *optical illusions*, which do not necessarily occur in a computer vision model, but should be apparent in a vision model that claimed to represent the way human vision works, or a vision system that tries to identify the same patterns and features that a human would. This area of research leads to a shibboleth for testing *bioplausible* models of vision.

Bioplausible models must satisfy two criteria, *computational feasibility* and *neurological plausibility*. Although in vision generally, and in the perception of optical illusion in particular, there are many levels of processing involved from eye to cortex before the final perception of the visual scene or illusory pattern, we are particularly interested in the underlying neural activity of retinal/cortical lower level processing. The sombrero-like interaction model we explore (Fig. 1) is prevalent through these early levels and, according to our model, it is here that the first cues to tilt emerge in geometric illusions (Fig. 2) such as in the Café Wall (Figs 2-5) where mortar lines appear to diverge and converge.

The model we are using here is an *ON-center receptive field (RF) model*, implementing retinal cell responses to the characteristics of visual scene. Although similar bioplausible models for implementing the response of retinal Ganglion Cells (GCs) to the Café Wall stimulus have been proposed by others [1, 2], none have quantified the degree of tilt by computational analysis. Also a systematic approach to *multiscale analysis* of the model outputs is missing in the explanations of previous studies although the effects of scale are illustrated in [2, 3]. Some experiments have been reported where a proposed model explaining the illusion was tested psychophysical experiments on human subjects [4, 5]. Many of these theories/explanations, however, remain at a descriptive level [6, 7], with little consideration of the underlying neurological mechanisms involved in the emergence of tilt illusion.

There is a reasonable understanding of illusions that depend on colour, brightness and contrast effects [8, 9], but there is a lack of general explanatory model for *tilt/tile illusions* like Café Wall patterns, and the Bulge illusions resulting from superimposed dots on top of a checkerboard background [3]. Ninio (2014) in his comprehensive study of geometric illusion explanations noted that the twisted cord family of illusion could not be explained by the proposed 'orthogonal expansion' and 'convexity principle' [10] which are the most general explanations for the many geometric/tilt illusions he described (Café Wall is from twisted cord family). So tile illusions remain an open area of research.

Many geometric illusions have a *highly directional effect* as in Café Wall. Their explanation for misperceived tilt typically depends on physiological interpretations of orientation detectors in the cortex and the lateral inhibition of these detectors. In the Café Wall, the *emergence of slanted line segments* along the mortar lines is claimed [1, 2] to be the reason for tilt percept in the pattern. These line segments result in the appearance of tiles as *wedge-shaped* [6] in a *local* view, which leads to a perception of alternating converging and diverging mortar lines at a more *global* level. Many theories for the Café Wall illusion involve high level explanations such as *'Border locking'* [6] and *'phenomenal model'* [7], and others have low level explanations such as *'brightness assimilation and contrast'* [11] and *'band-pass spatial frequency'* [1, 2].

Westheimer [12] offers a hybrid retino-cortical explanation for Café Wall illusion that considers multiple neural/neuronal processing stages involved in the tilt effect, claiming that the *irradiation inducing* effect, the apparent enlargement of white elements, cannot explain the Café Wall illusion and that additional stages of retinal/cortical processing need to be considered in any explanation. His retinal processing stages include *light spread*, *compressive nonlinearity*, and *center-surround transformation*. His cortical stages involve sharp straight borders, pointed corners and angle shifts in the final perception of the stimulus for illusion explanation.

It may seem that the explanations cited here explain the illusion at different levels of processing, but at a deeper level they have common features involving the innate neural mechanisms of lateral inhibition and recurrent suppression [13, 14, 15] of *retinal* and *cortical cells*.

In the micro-anatomy of the visual retinal receptive field (RFs), first the visual signal from the photoreceptors (rods and cones) is passed to bipolar cells and then to retinal ganglion cells (RGCs) whose axons carry the visual signal to the cortex. There are also two types of interneurons, providing lateral interaction with other elements, which called horizontal and amacrine cells. These latter cells receive input from numerous cells because of large dendritic arbors. Only one exception to the rule of lateral interaction is known: the innermost area of the fovea, with a high special resolution to a 1:1 relationship with direct throughput between photoreceptors, bipolar cells, and ganglion cells [16].

Recent retinal physiological findings have deepened our understanding of RGCs and their functionality. A multiscale representation and processing in the visual cortex of mammals and in the retina have been supported by physiological and psychophysical findings [17, 18, 19]. In a comprehensive study about retinal circuitry and coding, Field and Chichilnisky [18] reported the existence of at least 17 distinct RGC types inside the retina each with their definite encoding role. The variations of RFs type and their size change due to the eccentricity (the distance from the fovea) as well as intra-retinal circuitry, all indicating the mechanism of multiscale encoding inside the retina. Earlier, it was assumed that orientation detection takes place solely in the cortex, but it is found that some retinal cells have an orientation selectivity property similar to the cortical cells [18, 19], consistent with the raw-to-full primal sketch of Marr's theory of vision [20, 21].

The center–surround receptive field organization in RGCs is commonly believed to be due to lateral inhibition (LI) in the outer retina and the inner retina [22]. At the first synaptic level, the mechanism of lateral inhibition [13] enhances the nerve synaptic signal of photoreceptors, where activated cells inhibit the activations of nearby cells. This retinal neuronal processing is specified as a retinal pulse response or point spread function (PSF) that is a biological convolution with the effect of edge enhancement [15]. It acts as a bandpass filter that facilitates vision tasks. Inhibition at the second synaptic level (in the inner retina) is thought to mediate more complex response properties such as directional selectivity [22]. The model here is using the contrast sensitivity of RGCs based on a circular center and surround organization for the retinal RFs [3, 23, 24].

Eye movements like gaze shifts, fixation and pursuit, all affect our perception, since they carry the image across the retinal photoreceptors. Even in the 'Fixational eye movements' [25] there is a critical mechanism to prevent fading of the whole visual world involving a continuous shifting of retinal image by factor of a few 10s to 100s GCs on the retina due to the type of 'fixational eye movements'. This includes Microsaccades, Tremors and Drifts, which are unconscious source of eye movements. There are other intentional and unintentional eye movements while we look at the scene (pattern), notably full saccades and gaze shifts, which allow the high resolution fovea to rapidly scan the field of vision for pertinent information. All of these effective movements of the retinal image mean that retinal RFs around a point have different sizes depending on where the fovea is centered, so that different parts of the visual scene are processed at different scales at different times.

In Part II of this paper we (A) develop a simple Difference of Gaussian (DoG) model as a basic bioplausible model of lateral inhibition, (B) introduce the range of parameters and processing steps we will use in our analysis, (C) show that our restriction to consideration of ON-center OFF-surrounds cells and mid-level mortar shade is appropriate, and (D) summarize the entire modeling and analysis pipeline used in our experiments. We then present results for two experiments in Part III, (A) investigation of local 'cropped' samples, simulating foveal locus only, and (B) investigation of the entire image, simulating peripheral awareness and Gestalt processing.

## II. MODEL

### A. Retinal bioplausible model

Physiological evidence [18, 19] shows a diverse range of receptive fields with varying sizes in the retina, with size being a function of the type and eccentricity of the cell [26]. These suggest a multiscale retinal encoding [27] of the visual scene with the adaptation of retinal receptive fields (RFs) to textural elements [15, 28].

The history of the receptive field models back to Kuffler's demonstration of roughly concentric excitatory center and inhibitory surround [29]. It have been showed by Rodieck and Stone [23] and Enroth-Cugell and Robson [24] that the signals from the center and surround regions of photoreceptor outputs can be modelled by two concentric Gaussians with different radii [30, 31]. The computational studies and modeling of early visual processing were followed by Marr and Ullman [32] who were inspired by Hubel and Wiesel's [17] discovery of directional sensitive simple cells in the primate visual cortex. Marr and Hildreth [20] proposed Laplacian of Gaussian (LoG) as the optimal operator for initial filtering of retinal cells and noted that it can be approximated by a difference of Gaussians (DoG) with a ratio of diameters of ~1.6. A classical receptive field (CRF) model implementing retinal GCs responses [3] is used here to explain the emergence of tilt in Café Wall illusion, provide quantitative measurement of tilt angle in the pattern, and explain how incorporation of different foveal points during saccade lead to different tilt or bulge phenomena due to their perception at different scales and their integration into a multiscale map.

## B. Formal description and parameters

The main stage in our experiment generates a bioplausible representation for the image which is interpretable as the image edge map using the DoG model. This feature representation of the edges clearly reflects the perceived tilt in the image. For a sample cropped section of a Café Wall image, its feature map with multiple scales of DoG is shown in Fig. 2, in a binary form as well as false colored using the jetwhite[1] color map.

Applying Gaussian filter on an image generates a smoothed/blurred version of the image. The DoG output of an image is the difference between two blurred versions of the image, which is similar to band pass filtering. The easiest way of calculating the DoG output of an image is to generate the DoG filter first and then apply the filter on the image (one convolution). Fig. 1 illustrates 2D representations of two separate Gaussian filters for Center and Surround with their difference giving the DoG filter.

The DoG output of the retinal GCs model with the center and surround organization for a 2D image such as *I*, is given by:

$$\Gamma_{\sigma,s\sigma}(x,y) = I \times 1/2\pi\sigma^2 \, exp[-(x^2+y^2)/(2\sigma^2)] - \qquad (1)$$
$$I \times 1/2\pi s^2\sigma^2 \, exp[-(x^2+y^2)/(2s^2\sigma^2)]$$
$$s = \sigma_{surround}/\sigma_{center} = \sigma_s/\sigma_c \qquad (2)$$

where *x* and *y* characterize the distance from the origin in the horizontal and vertical axes respectively and $\sigma$ refers to $\sigma_c$, the sigma of the center Gaussian. The sigma of the surround Gaussian is represented by $\sigma_s=s\sigma$. Parameter *s* is referred to as the *surround ratio* here. The concentric representation of the center and surround Gaussians models the retinal point spread function (PSF) and retinal lateral inhibition (LI) [23, 24, 30].

The second derivative of the Gaussian can be estimated as the difference of two DoGs and is referred to as the Laplacian of the Gaussian (LoG). It has been shown that for modeling the RFs of retinal GCs, DoG [30, 31] is a good approximation of LoG when the ratio of dispersion of center to surround, $s \approx 1.6$ ($\approx \varphi$, the ubiquitous Golden Ratio) [20]. Increasing *s* leads to surround suppression covering a wider area while its height declines. In the experimental runs reported in this paper, $s = 2$ is used for convenience (other commonly used values like $1.4 \approx \sqrt{2}$ and $1.6 \approx \varphi$ show little difference).

As a further practical matter, it is inconvenient to deal with Gaussians of unbounded extent, so the DoG model is only applied within a window of a size chosen so that the value of both Gaussians is insignificant outside the window (less than 5% for the surround Gaussian). We therefore control *windowSize* as large windows have high computational cost. The *windowSize* is determined by the *h* (*window ratio*) parameter and $\sigma_c$ as shown below:

$$windowSize = h \times \sigma_c + 1 \qquad (3)$$

Parameter *h* determines how much of the center and surround Gaussians are included in the filter, and in this paper we standardize on $h = 8$ (two standard deviations of surround).

---
[1] Downloadable from MathWorks central file exchange. http://www.mathworks.com/matlabcentral/fileexchange/48419-jetwhite-colours-/content/jetwhite.m

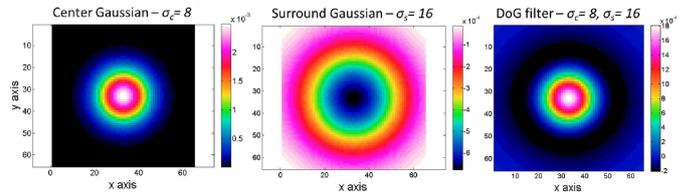

Fig. 1. Left: Center Gaussian with $\sigma_c = 8$. Center: Surround Gaussian with $\sigma_s = 16$ (surround ratio $s=2$). Right: The Difference of Gaussian result. Represented in jetwhite color map. (Reproduced by permission from [36])

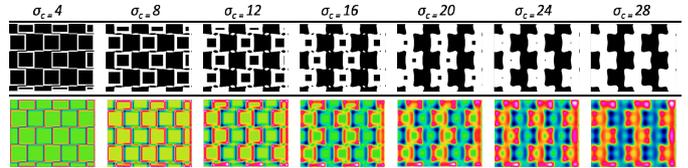

Fig. 2. Top: Binary edge map for a Café Wall pattern with 200×200px Tiles (*T*) and 8px Mortar thickness (*M*). Bottom: jetwhite color code for the edge map. $\sigma_c$ ranges from 4 to 28 with incremental step of 4 with *surround ratio s = 2, and window ratio h = 8*. (Reproduced by permission from [36])

## C. ON- and OFF- cell responses to Café Wall stimulus

The biological function of ON- and OFF- cell responses encodes visual objects in terms of their deviation from mean luminance. This is done by two sets of neurons, ON-cells for light increment, and OFF-cells for light decrement [16, 33].

The tilt illusion in the Café wall pattern seems to be the result of the appearance of the small slanted line segments on the mortar lines that connect two same colored tiles in two adjacent rows in the pattern. These slanted line segments are sometimes referred to as the twisted cord elements [1, 2, 3, 5] inside the pattern. The sensitivity profile and circularly symmetric activity of the ON-center and OFF-center ganglion cells creates peaks and troughs, which generate these elements as their processing output of the stimulus.

The color of the mortar line should be a gray in the intermediate range between black and white tiles [1, 6]. The reason is that an illusory tilted segment from a white tile to a white tile is perceived when the mortar is lighter than the two black tiles on each side, due to the OFF-center detectors, but the corresponding tilted segment from a black tile to a black tile is perceived when the mortar is darker than the two white tiles on each side. For both effects to happen with equal strength requires roughly equal distance between the mortar shade and the black and white shades.

The stimulus in Fig. 3 is a Café Wall 8×12 pattern with 50×50px Tiles and 2px Mortar thickness. For easier notation we recall the pattern Café Wall 8×12-T50-M2 in which 8×12 shows #rows and #columns of Tiles in the pattern, *T* stands for TileSize and *M* for MortarSize followed by their values.

In both ON- and OFF- cells implementations, the model parameters are $\sigma_c=3$, *surround ratio s=2*, and *window ratio h=8*. The ON-center and OFF-center RFs responses on the given Café Wall pattern, Fig. 3 (Left), are presented in the Center and Right of the Figure. The output of their response indicates that both ON- and OFF-cell responses to the stimulus, highlight the same direction of convergence and divergence in the tilt effect. The only difference in the output result is that every one of them clearly highlights the twisted cords between the two same colored tiles due to the connection of mortar lines and the ON-center and OFF-center activities noted before. Therefore in the modeling of RFs for tilt detection in the Café Wall pattern there is no difference which cell type is implemented. Both provide similar tilt orientation.

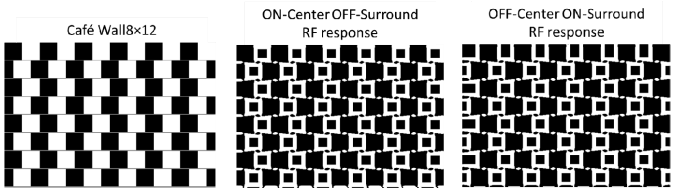

Fig. 3. Left: Café wall 8×12 with 50×50px Tiles and 2px Mortar. Center: ON-center OFF-surround RF response. Right: OFF-center ON-surround RF response – $\sigma_c=3$, surround ratio $s = 2$, and window ratio $h = 8$.

### D. Model and processing pipeline

The DoG transformation induces the tilted line segments in the pattern, and quantitative measurement of tilt angles let us to compare them with the tilt perceived by a human observer. For this we embed the DoG model in a processing pipeline involving multiple standard image processing transformations.

#### 1) MODEL

The sigma of the center Gaussian ($\sigma_c$) is the central parameter of the model and its optimal value is dependent on the dimensions of the basic tile and mortar elements. To extract the tilted line segments along the mortar lines, $\sigma_c$ should be similar in size to the Mortar thickness for mortar related edges to be detected. Fig. 2 shows the output of the MODEL for a cropped section of a Café Wall pattern with 200×200px Tiles (T) and 8px Mortar thickness (M). Based on the stimulus pattern and the fixed surround ratio and window ratio, (relative to $\sigma_c$) we use a range of 0.5M to 3.5M, with incremental step of 0.5M for $\sigma_c$. So the output of the MODEL is a form of edge map with multiple scales of DoG. Now we illustrate how to measure the slope of the detected tilts in the edge map.

#### 2) EDGES

The MODEL is applied at multiple scales of DoG (with the same surround and window ratios) providing a form of edge map. Then we measured tilt angles as follows: At each scale, first the edge map is binarised and then Hough Transform (HT) [34] is applied to measure the tilt angles in detected slanted line segments in the edge map. We need the slope of these lines, so the Hough representation is used here. HT uses a two-dimensional array called the *accumulator* to store lines information with quantized values of $\rho$ and $\theta$ where $\theta$ is in the range of $[0, \pi)$. Every edge pixel $(x, y)$ in the image space, corresponds to a sinusoidal curve in $(\rho, \theta)$ space (Hough space) as given by (4):

$$\rho = x.\cos\theta + y.\sin\theta \qquad (4)$$

where $\rho$ indicates the distance between the line passing through that point with a specific $\theta$ identifies the origin, and $\theta$ is the counter-clockwise angle between the normal vector ($\rho$) and the positive direction of the x-axis.

The output of the Hough transform is a two-dimensional *accumulator matrix H*, with the dimension of $\rho \times \theta$. Each element of the matrix corresponds to the number of pixels located on the line represented by quantized parameters of $(\rho_i, \theta_j)$. So the output of the EDGES is H matrix representing the edge map in Hough space.

#### 3) HOUGH

All possible lines that could pass through every edge point in the edge map are extracted in the EDGES processing stage, but we are more interested in the detection of tilt induction line segments inside the Café Wall pattern. Two MATLAB functions called houghpeaks and houghlines have been used here for this reason. The local maxima in the accumulator space (H) show the most likely lines that can be extracted.

The 'houghpeaks' function finds the peaks in the Hough accumulator matrix H, having the parameters of NumPeaks (maximum number of lines to be detected), Threshold (threshold value for searching H for the peaks), and NHoodSize (the neighborhood suppression size which set to zero after the peak is identified). The 'houghlines' function extracts line segments associated with a particular bin in a Hough accumulator matrix. Its parameters are "FillGap" (the max gap allowed between two line segments associated with the same Hough bin, which result in merging them to a single line segment) and "MinLength" (the min length for merged lines to be kept).

Fig. 4 illustrates a sample output of the HOUGH analysis stage. The investigated pattern is a crop section of a Café Wall 9×14-T200-M8 (Fig. 5-Left). The detected line segments are shown in green, displayed on a binarized edge map with seven DoG scales of 4, 8, 12, 16, 20, 24, and 28. Blue lines indicate the longest detected line segment.

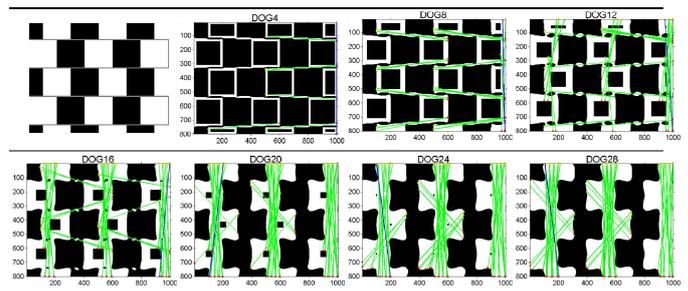

Fig. 4. HOUGH stage result on a cropped section of a Café Wall with 200×200px Tiles and 8px Mortar. Detected Hough lines are drawn on a seven scales edge map of the stimulus. $\sigma_c$ ranges from 4 to 28 (DoG4 : $\sigma_c=4$).

#### 4) ANALYSIS

Four reference orientations are defined including horizontal (H), vertical (V), positive diagonal (+45º, D1), and negative diagonal (-45º, D2), and an interval of [-22.5º, 22.5º) around them is chosen to cover the whole space. So the information of the detected line segments from HOUGH are saved inside four matrices based on how close they are to one of these reference orientations for tilt analysis. As shown in Fig. 4 in fine scales, near horizontal lines are detected but as the scale of DoG increases, the mortar lines are disappeared, and the *near horizontal tilt* is replaced by *zigzag vertical* lines joining similar colored tiles [3]. The statistical analysis of the detected lines in the neighborhood of each reference orientation is the output of this stage which further explained in the experimental results section.

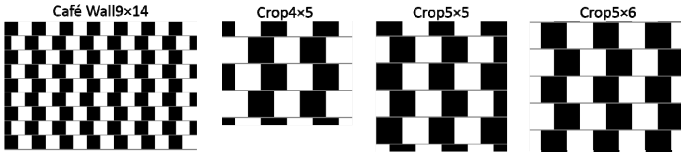

Fig. 5. Café Wall stimulus with 200×200px Tiles, and 8px Mortar with three "foveal" crop sizes explored. (CropH×W is a H×W Tile size area)

III. EXPERIMENTAL RESULTS

In this section, we report on two example experiments on evaluating the model and tilt analysis, correspond to local and global tilt and foveal/peripheral view of the pattern.

Our visual perception of tilt changes when we are fixating on a small section of the pattern. For instance, when we fixate on a part of a mortar line, the tilt in a close by region to our focusing point weakens, but we still have a peripheral tilt perception which results in maintaining the overall tilt recognition of the pattern. It seems that the peripheral tilt recognition has a higher impact on our final perception of the pattern compared to the weak tilt perception of some foveal/local focusing regions. This peripheral/global understanding provides a wholistic impression of the visual field, and can be linked to a Gestalt psychology percept of tilt induction patterns.

In the fovea, the acuity is high due to high density and small size receptors. As eccentricity increases, the acuity declines with increasing RF sizes and nearest neighbor distances. The model used here was inspired by the first proposed model for foveal retinal vision by Lindeberg and Florack [27]. The model is based on *simultaneous* sampling of the image at all scales, and since our vision is scale-invariant, so what is sent to the brain, is not a single image, but a *stack* of images, a *scale-space*. All scales are separately and near independently encoded from the incoming intensity distribution [28].

A. Experiment 1

The aims of this experiment are concluded in the evaluation of the effect of sampling size on detected mean tilt value of the Café Wall and the possible correlations to the foveal/peripheral view of the pattern due to gaze shifts and saccades. The investigation uses local 'cropped' samples for simulating foveal-sized locus only (but different scales occur when saccadically shifted to different degrees of eccentricity in the periphery).

We confine consideration initially to the pattern Café Wall 9×14 with 200×200px Tiles and 8px Mortar (Fig. 5), and fix parameters not being investigated at this stage. Three "foveal" crop sizes are explored in this experiment: Crop4×5 (Crop window size of 4×5 Tile size area), Crop5×5, and Crop5×6, of which an example for each sample group is given in Fig. 5. The size of foveal image can be estimated based on the pattern's visual angle and the size of the fovea which is approximately 0.01 mm$^2$ (20 arcmin of visual angle - 20×20px), but the sample sizes are selected here for convenience without considering human subject, particular image size, or viewing distance.

For each specified crop window size, 50 samples are taken from Café Wall 9×14 in which for the first sample, the top left corner is selected randomly from the pattern, and then for the rest of samples, there is a horizontal shift of the cropping window with an offset of 4 pixels between samples. This covers a total shift of a Tile size (200px) after the sampling is finished, and guarantees no repetition of samples in each set.

As described in previous section, for the edge map representation of the samples, the $\sigma_c$ parameter is chosen in the range of *0.5M* to *3.5M* with incremental step of *0.5M*. Finer scales are unhelpful due to the Mortar size of 8px, and coarser scales exceeding the Tile size, similarly result in very distorted and uninteresting edge pattern. Viz. the DoG scales in the model should be of the same order as the features we are interested in capturing in the pattern. One of the main advantages of the scale-space bioplausible model used is that the result is not very sensitive to specific characteristics of the pattern elements. Only an initial adjustment for $\sigma_c$ range is needed with its incremental step.

The parameters of *houghpeaks* and *houghlines* functions should be selected in a way to detect the slanted line segments in the pattern in lower scales. E.g. *MinLength* should be larger than *TileSize* to avoid the detection of the outlines of the tiles, and *FillGap* value should fill small gaps between line segments appear on the edge map from Mortar lines and Tiles borders at lower scales to detect near horizontal tilted lines. These parameters have been chosen empirically based on the pattern's characteristics, and kept constant for all the experiments. *Numpeaks*=100, *Threshold*=3, *FillGap*=40, and *MinLength*=450.

The results of mean tilt for each sample set is box plotted in Fig. 6 for 4 reference orientations of Horizontal (*H*), Vertical (*V*), and Diagonal (*D1*, *D2*), and at 7 DoG scales of their edge map.

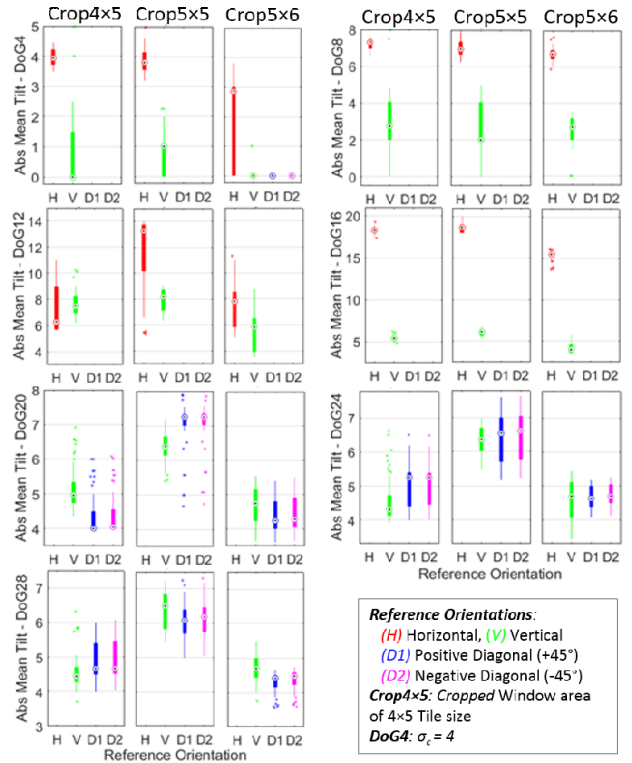

Fig. 6. Boxplot of mean tilt for 3 "foveal" crop sizes explored for 4 reference orientations of (H, V, D1, D2), at 7 DoG scales ($\sigma_c$=4, 8, 12, 16, 20, 24, 28), s=2, h=8, and Numpeaks=100, Threshold=3, FillGap=40, and MinLength=450. (Reproduced by permission from [36])

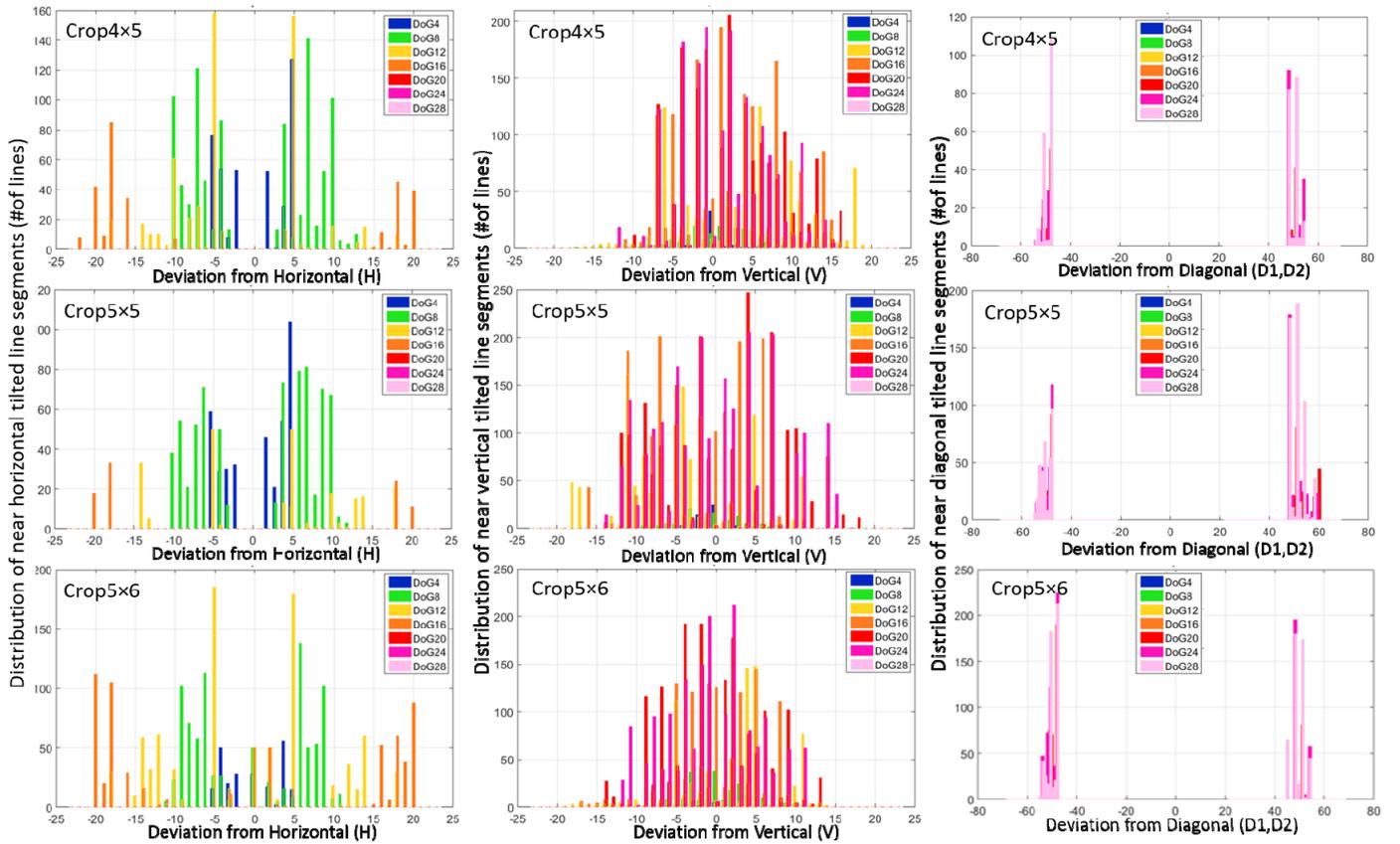

Fig. 7. The distribution of near Horizontal (Left Column), near Vertical (Center Column), and near Diagonal (Right Column) detected line segments of 3 "foveal" crop sizes for 7 different DoG scales. The other parameters are the same as Fig. 6. (Reproduced by permission from [36])

As Fig. 6 indicates, in the first 4 scales, only horizontal and vertical lines are detected. The horizontal tilted line segments which induce the tilt illusion appear in these scales. Among these 4 scales, DoG8, which matches with the Mortar size detects the horizontal tilt in a nearly stable range around 7° in all samples. As the scale increases from 20, more vertical and diagonal lines are extracted while no other horizontal lines are detected. This is due to the enlargement of the outlines of the tiles by increasing $\sigma_c$, which results in more line detection in the coarse edge map scales. The results show that for Horizontal mean tilt deviation, as the scale of the DoG model increases, the mean tilt value also increases, although it is around 8° for DoG8 and 12. In the finest scale (DoG4) however, the horizontal tilt angle is quite small (3-4°) compared to DoG8. This suggests why we perceive the tilt effect in the pattern in much weaker magnitude when we are fixating on the pattern, since similarly in the fovea the acuity is high due to high density and small size receptors. For the vertical deviation, the mean tilt has a slight increase at the first two scales and nearly stays around 5° from V reference orientation (axis). In the diagonal tilt investigation, the mean tilt deviation is around 4° and 5° from *D1*, and *D2* axes which can be seen after DoG20.

Comparing the results at a given scale, the detected tilts show slight differences across sample sets, and this is to be expected given the random sampling and the fixed parameters relating to *houghpeaks* and *houghlines* which are kept constant here rather than optimized for each scale. In particular we can expects edge effects to interact with both the random sampling and the *Numpeaks* parameter, which are related to the sample sizes, but kept constant here for consistency of the higher level analysis/model. The parameters chosen here highlight the dominant lines and their orientations we are most interested in. The tilt detection results are reliable when compared to our angular tilt perception of the pattern, while the computational cost of the model and tilt analysis is reasonable. However, the parameter values of *houghpeaks* and *houghlines* could be optimised later for more accurate results.

The mean tilt results for the detected lines for all 50 samples in each set are further analysed across all the DoG scales. The investigations show that for horizontal orientation, there is an increase in mean tilt as well as the standard deviation from the mean by $\sigma_c$ increases. Vertical and Diagonal lines are started to be detected as the edges thicken to the same order as the tiles at coarse scales, and this correlates to different ways of grouping tiles in the Café wall pattern at different scales [3].

Fig. 7 shows the distribution of lines near each reference orientation (*H, V, D1, D2*) for 3 sample sets by considering every DoG scales for easier comparison. The results of the near diagonal tilted lines have been graphed together for fairer representation and comparison versus the alternating up and down tilting horizontals and the zigzagging verticals. All the graphs given indicate the effect of the edge map scale on the range of detected tilts. This range covers a wider area around reference orientations when the DoG scale increases. Also the number of detected lines is highly dependent with the size of sample.

In Fig. 7 (Left-column), the detected near horizontal lines are given for three sample sets. In the scale of 16, there is a high range of variations of tilt angle that is not reflected in our subjective perception of the pattern. Furthermore, in scale 4 the angular values of detected slanted lines are very small. So based on these results, the most informative parameter for the DoG scale in order to detect the convergence and divergence of the mortar lines, is a center scale near the size of the Mortar, here DoG8. Also the color codes based on the legend highlighted the fact that horizontal lines are detectable in fine scales, when there is still some parts of Mortar lines are left in the DoG output.

In Fig. 7 (Center-column) the detected near vertical lines are given, and although as Fig. 6 indicates, they start to be detected in fine scales, but the majority of vertical lines are in DoG20, and 24 (Look at the edge effect of vertical lines in DoG8 and 12 in Fig. 4). In Fig. 7 (Right-column) the distributions of the detected near diagonal lines are graphed with their deviations from *D1*, and *D2* axes in the same graph for each sample set. The graphs indicate their detections are mainly around the coarse scales of DoG24, and 28.

In [3] the Café Wall illusion explained based on incompatible grouping of tiles in lower scales and in higher scales, in which our result is a proof for that. In low scales the tiles are connected by the Mortar lines in horizontal direction (slanted line segments - twisted cord elements) and in higher scales, when the Mortar lines have been disappeared, the zigzag vertical grouping of tiles is happening in an opposite direction. We claim that this is the main reason of the tilt illusion in the Café Wall pattern. This view has been considered both local/global tilt effects of the pattern, as well as a simple way of explanation of the visual angle and the viewing distance to the pattern. This distance results in our perception of the pattern due to our recognition of what pattern elements are observable from near and from far, which is captured here as multiple map representations at different DoG scales.

### B. Experiment 2

The aims of this experiment are to confirm the robustness of the model in local and global tilt analysis of the Café Wall illusion by comparison of the tilt results of "foveal" size sample sets with the "peripheral" tilts of the whole pattern. We investigate the Gestalt pattern, simulating peripheral awareness across the entire image in here.

For this the Café Wall 9×14 pattern with 200×200px Tiles and 8px Mortar have been investigated for quantitative measurement of tilt in 4 reference orientations. The results of the investigation are shown in Fig. 8, including error bars to provide an indication of significance.

As the horizontal tilt deviation shows, the mean tilt value increases as the edge map scale increases, and we have nearly similar tilt value around the scale of 8 compared to sample set results in Fig. 6. Vertical tilt detection in higher scales is around 5 ° in the sample set results, and here around 2°, while the diagonal deviation range is approximately 3° here while it was about 1º degree more in the crop sample sets. We discuss these results further in the conclusions.

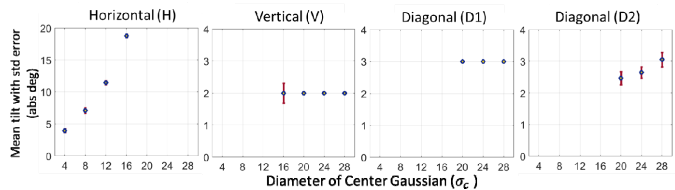

Fig. 8. Mean tilt and standard error arround reference orientations (*H, V, D1, D2*) for the Café Wall9×14 pattern with 200px Tiles and 8px Mortar, displayed for every DoG scales. (Reproduced with permission from [36])

Fig. 9 indicates the detected tilted lines in DoG scale of 16 ($\sigma_c=16$) for the Café Wall 9×14, as well as one sample of each foveal set. The Hough parameters kept exactly the same, for all the samples and analysis of the whole pattern: *Numpeaks=100-Threshold=3-FillGap40-MinLenght450*.

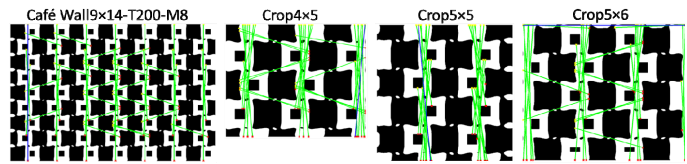

Fig. 9. *HOUGH* stage result of a Café Wall with 200px Tiles, and 8px Mortar (Left) for DoG scale of 16 ($\sigma_c=16$). Hough line results on 3"foveal" cropped window samples drawn on the scale of $\sigma_c=16$ of their edge map (3 crops on the Right).

## IV. CONCLUSIONS AND FUTURE WORK

A bioplausible model implementing ON-cells retinal receptive field response to the stimulus as Difference of Gaussian is used to generate a bioplausible intermediate representation at multiple scales that reflect differences in the dominant tilts apparent in the Café Wall illusion. We argued that the edge/scale information from both the fovea and the periphery, as edge maps for detection of features at different scales, are being combined in the cortex for our final percept. Thus different scales are represented in fovea versus periphery and at different distances and image sizes, explaining the illusory changes.

Although the perceptual effect is highly directional in the pattern, our model illustrates how the lateral inhibition and suppression effects in the retinal/cortical simple cells, are responsible for the emergence of tilt in the pattern, even though for the angle detection of tilt, further processing by orientation selective cells in the retina/cortex might be needed. In here for the detection of tilt angle, we have exploited an image processing pipeline to detect the degree of mean tilt and distributions of tilted line segments around 4 reference orientations of Horizontal (*H*), Vertical (*V*), and Diagonal (*D1, D2*) covering the whole space.

The experimental results on 3 foveal crop sample sets and the whole Café wall pattern shows that, the model is capable of tilt detection, and the result are nearly consistent in the sample sets while we kept all the parameters the same during the experiments on every input image. One of the reasons for the minor differences of tilt in the global tilt analysis is due to the parameter value of *Numpeaks*, which is kept constant across the cropped samples as well as whole pattern, which is around

5 times of the size of the cropped (foveal) samples. The density of detected lines in the edge map shows the effect clearly in Fig. 9, but increasing *Numpeaks* with size reduces the effect.

No psychophysical tests have been performed to validate the predictions implicit in our results, and this is one of our future research priorities. Furthermore, one of the effects a viewer of a Café Wall pattern notices is that the tilts seem larger in their peripheral vision than at their focal point, as predicted by our quantified results: larger DoGs corresponding to the lower resolution of the periphery gives rise to larger perceived angles in both experiments, before losing the mortar lines completely and seeing zigzag vertical patterns or eventually diagonal patterns similar to the brightness illusions (corresponding to viewing from greater distances). A related aspect for future work is to explore relation between DoG, mortar and tile size, and visual angle, and make predictions for different distances and apparent size.

Extension of the model to non-CRFs (nCRF) [35] based on elongated surrounds, might also facilitate the directionality evaluation of the tilt in tile illusions.

ACKNOWLEDGMENT

Nasim Nematzadeh was supported by an Australian Postgraduate Award for her PhD [36].